\documentclass[journal abbreviation, manuscript]{copernicus}
\usepackage{comment}
\usepackage{caption}
\usepackage{subfigure}

\newcommand{\EE}{\mathbb{E}}

\newcommand{\RR}{\mathbb{R}}

\newcommand{\One}{\mathbf{1}}
\newcommand{\intd}{\, \text{d}}
\newcommand{\vm}[1]{\ensuremath{\mathbf{#1}}}
\newcommand{\vms}[1]{\ensuremath{\boldsymbol{#1}}}

\newcommand{\transpose}{\text{T}}
\newcommand{\eexp}[1]{ \text{e}^{#1} }
\newcommand{\eps}{\varepsilon}
\newcommand{\pdf}{\text{pdf}}

\newcommand{\up}{\vm p} 
\newcommand{\rp}{r}  
\newcommand{\Nmc}{N_{\text{MC}}} 
\newcommand{\QoI}{Q} 
\newcommand{\dom}{D} 
\newcommand{\SD}{\Omega_{\text{s}}} 
\newcommand{\IRPo}{T} 
\newcommand{\IRP}{T_{\rp}} 
\newcommand{\dIRP}{T_{d}} 
\newcommand{\freq}{\omega} 

\usepackage{glossaries}
\newacronym{pde}{PDE}{partial differential equation}
\newacronym{gpr}{GPR}{Gaussian Process Regression}
\newacronym{gp}{GP}{Gaussian Process}
\newacronym{fem}{FEM}{finite element method}
\newacronym{fe}{FE}{finite element}
\newacronym{mc}{MC}{Monte Carlo}
\newacronym{pec}{PEC}{perfect electric conductor}

\newglossaryentry{qoi}
{
	name={QoI},
	description={quantity of interest},
	first={\glsentrydesc{qoi} (\glsentrytext{qoi})},
	plural={QoIs},
	descriptionplural={quantities of interest},
	firstplural={\glsentrydescplural{qoi} (\glsentryplural{qoi})}
}

\newcommand\change[1]{{\color{black}#1}} 

\begin{document}

\title{Yield Optimization using Hybrid Gaussian Process Regression and a Genetic Multi-Objective Approach}


\Author{Mona}{Fuhrländer}
\Author{Sebastian}{Schöps}

\affil{Technische Universität Darmstadt, Institute for Accelerator Science and Electromagnetic Fields and Centre for Computational Engineering, Darmstadt, Germany}




\correspondence{Mona Fuhrländer (mona.fuhrlaender@tu-darmstadt.de)}

\runningtitle{Yield Optimization using Hybrid-GPR and Genetic MOO}

\runningauthor{Fuhrländer and Schöps}

\received{}
\pubdiscuss{} 
\revised{}
\accepted{}
\published{}


\firstpage{1}

\maketitle

\begin{abstract}
Quantification and minimization of uncertainty is an important task in the design of electromagnetic devices, which comes with high computational effort. We propose a hybrid approach combining the reliability and accuracy of a Monte Carlo analysis with the efficiency of a surrogate model based on Gaussian Process Regression. We present two optimization approaches. An adaptive Newton-MC to reduce the impact of uncertainty and a genetic multi-objective approach to optimize performance and robustness at the same time. For a dielectrical waveguide, used as a benchmark problem, the proposed methods outperform classic approaches.
\end{abstract}


\introduction  
In the manufacturing process of electromagnetic devices, e.g. antennas or filters, uncertainties may lead to deviations in the design parameters, e.g. geometry or material parameters. This may lead to rejections due to malfunctioning. 
Therefore, it is of great interest to estimate the impact of the uncertainty before production is started, and to minimize it if necessary in the design process.

The yield is a measure for the impact of uncertainty, it is defined as the fraction of realizations in a manufacturing process fulfilling some defined requirements, the so-called performance feature specifications, cf.~\cite{Graeb_2007aa}. The relation between the yield $Y$ and the failure probability $F$ is given by $Y = 1 - F$. 
A well established method for estimating the yield is \gls{mc} analysis~\cite[Chap. 5]{Hammersley_1964aa}. 
Therefore, the performance feature specifications have to be evaluated on a large sample set.
In case of electrical engineering, very often this requires to solve partial differential equations originating from Maxwell's equations using the \gls{fem} for example. This implies high computational effort.

This work deals with the efficient estimation of the yield using a hybrid approach combining the efficiency of a surrogate model approach and the accuracy and reliability of a classic \gls{mc} analysis, cf.~\cite{Li_2010aa}. The surrogate model is based on \gls{gpr}, cf.~\cite{Rasmussen_2006aa},
which has the feature that the model can be easily updated during the estimation process, see~\cite{Fuhrlander_2020ab}. 
The optimization algorithm maximizing the yield is based on a globalized Newton method~\cite[Chap. 10.3]{Ulbrich_2012aa}. The presented method is similar to the adaptive Newton-MC method in~\cite{Fuhrlander_2020aa}, but \gls{gpr} is used here for the surrogate model instead of stochastic collocation. 
Due to the blackbox character of \gls{gpr}, the opportunity arises to use this algorithm in combination with commercial \gls{fem} software. Finally, we formulate a new optimization problem and propose a genetic multi-objective optimization (MOO) approach in order to maximize robustness, i.e., the yield, and performance simultaneously.

\section{The yield}

Let $\up$ be the vector of uncertain design parameters, which is Gaussian distributed, i.e., $\up \sim \mathcal{N}(\overline{\up}, \vms \Sigma)$, where $\overline{\up}$ denotes the mean value and $\vms \Sigma$ the covariance matrix. The probability density function of the multivariate Gaussian distribution is denoted by $\pdf(\up)$.
Following~\cite{Graeb_2007aa} the performance feature specifications are defined as inequalities in terms of one (or more) \glspl{qoi} $\QoI$, which have to be fulfilled for all so-called range parameter values $\rp$, e.g. frequencies, within a certain interval $\IRP$, i.e.,
\begin{equation}
\QoI_{\rp}(\up) \leq c \ \ \forall \rp \in \IRP.
\label{eq:pfs_general}
\end{equation}
Without loss of generality the performance feature specifications are defined as one upper bound with a constant $c\in\RR$.
The safe domain is the set of all design parameters, for which the performance feature specifications are fulfilled, i.e.,
\begin{equation}
\SD := \lbrace \up \,|\ \QoI_{\rp}(\up) \leq c \ \ \forall \rp \in \IRP \rbrace.
\label{eq:SafeDomain_general}
\end{equation}
Then, the yield is introduced as~\cite[Chap. 4.8.3, Eq. (137)]{Graeb_2007aa}
\begin{equation}
Y(\overline{\up}) := \EE [\One_{\SD}(\up)]
:= \int_{-\infty}^{\infty} \dots \int_{-\infty}^{\infty} \One_{\SD}(\up) \, \pdf(\up)  \intd \up,
\label{eq:Yield}
\end{equation}
where $\EE$ is the expected value and $\One_{\SD}(\up)$ the indicator function 
which has value $1$ if $\up$ lies in $\SD$ and $0$ otherwise.

\section{Yield estimation}

A classic \gls{mc} analysis would consist in generating a set of $\Nmc$ sample points and calculating the fraction of sample points $\up_i$ lying inside the safe domain. The \gls{mc} yield estimator is then given by~\cite[Chap. 5]{Hammersley_1964aa}
\begin{equation}
\tilde{Y}_{\text{MC}}(\overline{\up}) = \frac{1}{\Nmc} \sum_{i=1}^{\Nmc} \One_{\SD}(\up_i).
\label{eq:MCyieldEstim}
\end{equation}
A commonly used error indicator for the \gls{mc} analysis is 
\begin{equation}
\sigma_{\tilde{Y}_{\text{MC}}} = \sqrt{\frac{\tilde{Y}_{\text{MC}}(\overline{\up})(1-\tilde{Y}_{\text{MC}}(\overline{\up}))}{\Nmc}} \leq \frac{0.5}{\sqrt{\Nmc}},
\label{eq:stdYieldEst}
\end{equation}
where $\sigma_{\tilde{Y}}$ is the standard deviation of the yield estimator, cf.~\cite{Giles_2015aa}. 
Thus, for high accuracy a large sample size is needed, which means high computational effort, i.e., note the square root in the denominator. 

\begin{figure}%
	\centering
	\subfigure
	{\includegraphics[width=0.3\textwidth]{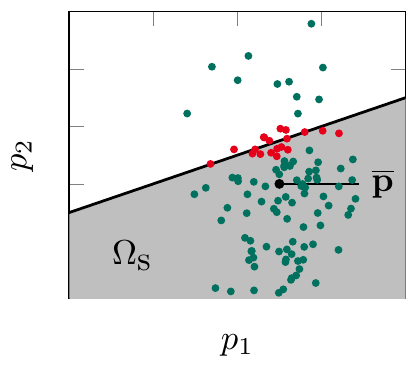}}%
	\qquad  \qquad
	\subfigure
	{\includegraphics[width=0.3\textwidth]{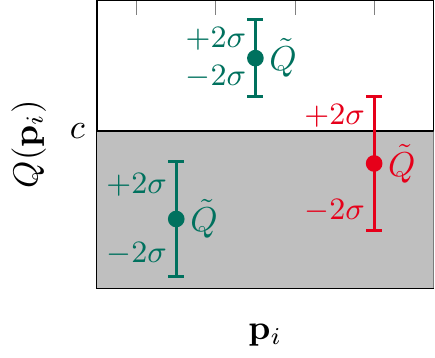}}
	\caption{Critical (red) and non-critical (green) sample points.}
	\label{fig:CriticalSamplePoints}
\end{figure}

There are different approaches in order to reduce the computational effort by either reducing the sample size, e.g. by Importance Sampling~\cite[Chap. 5.4]{Hammersley_1964aa}, or by reducing the computing time for one sample point, e.g. by using surrogate based approaches, see e.g.~\cite{Rao_1999aa,Rasmussen_2006aa,Babuska_2007aa}. 
In most surrogate approaches, the computational effort increases rapidly with an increasing number of uncertain parameters (curse of dimensionality), cf.~\cite{bellman1961curse}.
And, in~\cite{Li_2010aa} it is shown, that there are examples, where the yield estimator fails drastically, even though the surrogate model seems highly accurate.
The general idea of the hybrid approach proposed in~\cite{Li_2010aa} is to distinguish between critical sample points and non-critical sample points. Critical sample points are those which are close to the border between safe domain and failure domain,
see the red sample points in Fig.~\ref{fig:CriticalSamplePoints} (left) for a visualization with two uncertain parameters.
While most of the sample points are only evaluated on a surrogate model, the critical sample points are also evaluated on the original model. Only then it is decided if the sample point is accepted or not. By this strategy, accuracy and reliability can be maintained while computational effort can be reduced.
A crucial point is the definition of the critical sample points, i.e., of the term \textit{close to the border}. 

The hybrid approach we use in this work is based on~\cite{Li_2010aa} and is explained more detailed in~\cite{Fuhrlander_2020ab}. As surrogate model we use \gls{gpr}, i.e., we approximate the \gls{qoi} as \gls{gp} and assume that the error of the surrogate model is Gaussian distributed. For detailed information about \gls{gpr} we refer to~\cite[Chap. 2]{Rasmussen_2006aa}. 
One advantage of \gls{gpr} is that the standard deviation $\sigma(\up_i)$ of the \gls{gp} may serve as an error indicator of the surrogate model and thus can be used to find the critical sample points. With \gls{gpr} predictor $\tilde{Q}(\up_i)$ and safety factor $\gamma>1$ we define a sample point $\up_i$ as critical, if
\begin{equation}
\tilde{\QoI}_{\rp_j}(\up_i) - \gamma \sigma_{\rp_j}(\up_i) < c < \tilde{\QoI}_{\rp_j}(\up_i) + \gamma \sigma_{\rp_j}(\up_i)
\label{eq:CriticalSP}
\end{equation}
holds for any $\rp_j$, where $\rp_j$ is a discretization of $\IRP$ with $j=1,\dots,N_{\text{range}}$. In this case, $\QoI_{\rp_j}(\up_i)$ is evaluated on the original model, before classifying $\up_i$ as \textit{accepted} or \textit{not accepted},
see the red case in Fig.~\ref{fig:CriticalSamplePoints} (right, $\gamma=2$).
Then, the GPR-Hybrid yield estimator is calculated by
\begin{equation}
\tilde{Y}_{\text{GPR-H}}(\overline{\up}) = \frac{1}{\Nmc}\left( \sum_{\substack{\up_i \\ \text{crit.}}} \One_{\SD}(\up_i) + \sum_{\substack{\up_j \\\text{non-crit.}}} \One_{\tilde{\Omega}_{\text{s}}}(\up_j) \right),
\label{eq:YieldEstimGPRH}
\end{equation}
where $\tilde{\Omega}_{\text{s}}$ is the safe domain based on the performance feature specifications with the approximated \gls{qoi}.

Another advantage of \gls{gpr} is that we are not limited to specific training data, e.g. points on a tensor grid as in polynomial approaches, and thus we can add arbitrary points in order to update the \gls{gpr} model on the fly. This allows us to start with a rather small initial training data set and improve the \gls{gpr} model during the estimation by adding the critical sample points to the training data set without much extra cost.
Also, sorting the sample points before classification can increase the efficiency, cf.~\cite{bect2012sequential}.
For more details and possible modifications of the estimation with the GPR-Hybrid approach, we refer to~\cite{Fuhrlander_2020ab}.

\section{Yield optimization}

After estimating the impact of uncertainty, often a minimization of this impact is desired. 
In the following we propose two optimization approaches. 
In the first approach, we assume that a performance optimization was carried out in a previous step, and focus here on the maximization of the yield. The optimization problem reads
\begin{equation}
\max_{\overline{\up}} Y(\overline{\up}).
\label{eq:OptProb}
\end{equation}
The gradient and the Hessian of the yield can be derived in analytical form without any further knowledge nor computing effort, see~\cite[Chap. 7.1]{Graeb_2007aa}. This allows us to use a gradient based algorithm like a globalized Newton method \cite[Chap. 10.3]{Ulbrich_2012aa}.

From Eq.~\eqref{eq:stdYieldEst} we know that the size of the sample set is crucial for the accuracy of the yield estimator, but it is also crucial for the efficiency of the algorithm. Typically, in the first steps of the Newton method high accuracy of the yield estimator is not necessary as long as the gradient indicates an ascent direction. 
Thus, we modify the Newton algorithm from~\cite{Ulbrich_2012aa} and propose an adaptive Newton-MC method, where we start with a small sample size $\Nmc^{0}$. When the solution shows no improvement and the target accuracy $\hat{\sigma}_{\tilde{Y}_{\text{GPR-H}}}$ is not reached in the $k$-th step, i.e., 
\begin{equation}
\left| \tilde{Y}_{\text{GPR-H}}(\overline{\up}^{k-1}) - \tilde{Y}_{\text{GPR-H}}(\overline{\up}^{k}) \right| < \hat{\sigma}_{\tilde{Y}_{\text{GPR-H}}}
\quad \text{and} \quad
\frac{1}{2}\,
\hat{\sigma}_{\tilde{Y}_{\text{GPR-H}}} < \sigma_{\tilde{Y}_{\text{GPR-H}}(\overline{\up}^{k})},
\end{equation}
then the sample size is increased to $\Nmc^{k} = \Nmc^{k} + \Nmc^{0}$. 
The detailed procedure can be found in~\cite{Fuhrlander_2020aa}. As a novelty, here, the GPR-Hybrid approach is used for yield estimation.

In the second approach, we formulate a multi-objective optimization problem to optimize yield and performance at the same time. Without loss of generality it can be formulated as
\begin{align}
\min_{\overline{\up}} &f_m(\overline{\up}),  &&\hspace*{-5.5cm}m=1,\dots,M  \hspace*{5.5cm}\label{eq:MultiOptFormulation}\\
\quad \text{s.t. } &g_n(\overline{\up}) \leq 0, &&\hspace*{-5.5cm}n=1,\dots,N \notag \\
&\overline{p}_i^{\text{lb}} \leq \overline{p}_i \leq \overline{p}_i^{\text{ub}}, &&\hspace*{-5.5cm}i=1,\dots,\text{dim}(\overline{\up}), \notag
\end{align}
with $f_1(\overline{\up}) = -Y(\overline{\up})$ and $f_m(\overline{\up})$, $m=2,\dots,M$ are key performance indicators.
\change{By solving this optimization problem, we aim to obtain the pareto front, which is the set of pareto optimal solutions \cite[Def. 2.1]{Ehrgott_2005aa}. For all pareto optimal solutions holds, that one objective value can only be improved at the cost of another objective value. Classic approaches include the weighted sum or the $\varepsilon$-constraint method \cite[Chap. 3-4]{Ehrgott_2005aa}. However, these methods require convexity of the pareto front, which cannot be guaranteed in this context. Furthermore, they always find only one pareto optimal solution for each run of the solver, thus they need to be solved many times with different settings in order to approximate the entire pareto front. 
Genetic algorithms, however, approximate the entire pareto front in one run. But for this purpose, a swarm of solutions needs to be evaluated. Thus, genetic algorithms are also computationally expensive.
For more details about genetic algorithms we refer to~\cite{Audet_2017aa}, here we only outline the basic idea: }
After creating an initial population of solutions, the fitness (i.e. quality) of the individuals is determined according to their objective function values. By mutation and crossover (recombination) new generations with improved fitness are created and the best individuals are selected. 
\change{In contrast to the above mentioned methods, genetic algorithms do not require convexity or any other previous knowledge about the problem.}

\section{Numerical results}

From the time harmonic Maxwell's formulation on the domain $\dom \in \RR^3$ we derive the curl-curl equation of the E-field formulation
\begin{equation}
\nabla \times \left(\mu^{-1} \nabla \times \vm E_{\freq} \right)  - \freq^2 \eps \vm E_{\freq} = 0 \quad \text{on~}\dom.
\label{eq:E-field_strong_general}
\end{equation}
It is to be solved for the electric field phasor $\vm E_{\freq}$, where $\freq$ denotes the angular frequency, $\mu$ the permeability and $\eps$ the permittivity. Discretizing the weak formulation by (high order) Nédélec basis functions leads to an approximated E-field $\tilde{\vm E}_{\freq}$. 
The \gls{qoi} is the \gls{fem} approximation of the S-parameter
\begin{equation}
Q_{\freq}(\up) = q \left( \tilde{\vm E}_{\freq}(\up) \right).
\label{eq:QoI}
\end{equation} 

As benchmark problem we consider a simple dielectrical waveguide, see~\cite{Fuhrlander_2020aa}, with two uncertain geometrical parameters 
and two uncertain material parameters. 
For physical soundness we model them as truncated Gaussian distributed, cf. \cite{Cohen_2016aa}, with 
\begin{align}
\overline{\up} &= [10.36, 4.76, 0.58, 0.64]^{\transpose},\\
\vms{\Sigma} &= \text{diag}\left([0.7^2, 0.7^2, 0.3^2, 0.3^2]\right).
\end{align}
The geometrical parameters are truncated at $\pm 3$\,mm and the material parameters at $\pm 0.3$.
With the frequency as range parameter the performance feature specifications are
\begin{equation}
\QoI_{\freq}(\up) \leq -24\,\text{dB} \ \ \forall \freq \in \IRPo_{\freq} = \left[ 2\pi 6.5, 2\pi 7.5 \right] \text{ in GHz}.
\label{eq:pfs}
\end{equation}
The frequency range $\IRPo_{\freq}$ is parametrized in $11$ equidistant frequency points. Then, a sample point is accepted, if the inequality in Eq.~\eqref{eq:pfs} holds for all frequency points in the discretized range parameter interval $\dIRP \subset \IRPo_{\freq}$.
For the estimation we set
an initial training data set of $10$ sample points
and an allowed standard deviation of the yield $\hat{\sigma}_{\tilde{Y}_{\text{GPR-H}}} \leq 0.01$ which leads to a sample size of $\Nmc=2500$, cf. Eq.~\eqref{eq:stdYieldEst}. In order to build the \gls{gpr} surrogate the python package scikit-learn is used, cf.~\cite{scikit-learn}. 
For each frequency point, a separate \gls{gpr} model is built for the real part and the imaginary part. 
For the \gls{gp} we use the mean value of the training data evaluations as mean function and the squared exponential kernel 
\begin{equation}
k\left(\up,\up'\right) = \zeta \, \eexp{-\frac{\left|\up-\up'\right|^2}{2l^2}}.
\end{equation}
The two hyperparameters $\zeta \in \RR$ and $l>0$ are internally optimized within the scikit-learn package. For \gls{gpr} the default settings of scikit-learn have been used, except for the following modifications: the starting value for $\zeta$ has been set to $0.1$, its optimization bounds to $\left[10^{-5},10^{-1}\right]$ and the noise factor $\alpha = 10^{-5}$.

\begin{figure}%
	\centering
	\includegraphics[width=0.5\textwidth]{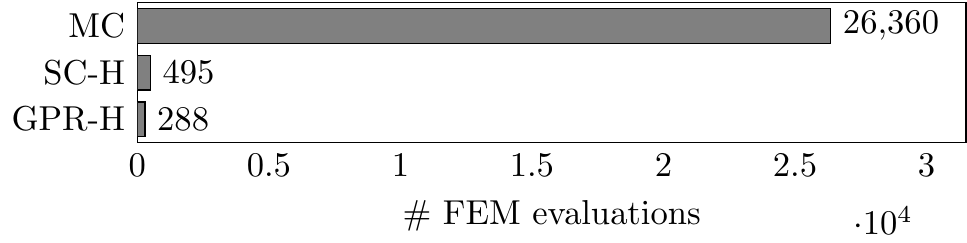}
	\caption{Computational effort for yield estimation.}
	\label{fig:CompYieldEst}
\end{figure}

We compare the proposed GPR-Hybrid approach with the hybrid approach based on stochastic collocation (SC-Hybrid) from~\cite{Fuhrlander_2020aa}, but without adaptive mesh refinement. As reference solution we consider the classic \gls{mc} analysis. Both hybrid methods achieve the same accuracy, i.e., the same yield estimator as the \gls{mc} reference, $\tilde{Y}_{\text{MC}}(\overline{\up}) = 95.44\,\%$.
In order to compare the computational effort we consider the number of \gls{fem} evaluations, necessary for solving Eq.~\eqref{eq:QoI}. In all methods a short circuit strategy has been used such that a sample point is not evaluated on a frequency point if it has already been rejected for another frequency point. Figure~\ref{fig:CompYieldEst} shows the comparison of these methods. The lower number of \gls{fem} evaluations in the GPR-Hybrid approach \change{compared to SC-Hybrid} can be explained by the fact that the surrogate model can be updated on the fly.

\begin{figure}%
	\centering
	\subfigure[Single-objective optimization: computational effort.\label{fig:YieldOpt_SingleEff}]
	{\includegraphics[scale=0.85]{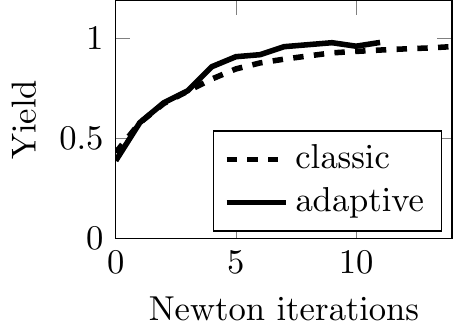}}%
	\qquad
	\subfigure[Single-objective optimization: yield progress.\label{fig:YieldOpt_SingleProg}]
	{\includegraphics[scale=0.85]{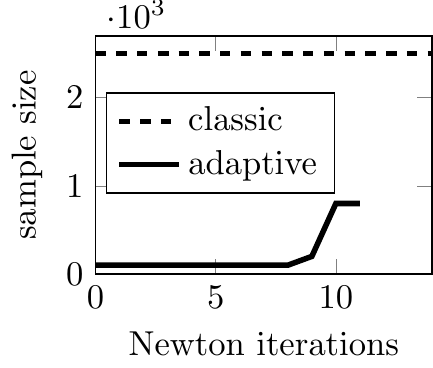}}
	\qquad
	\subfigure[Multi-objective optimization: pareto front for max (yield) and min (width) objectives.\label{fig:YieldOpt_MultiFront}]
	{\includegraphics[scale=0.85]{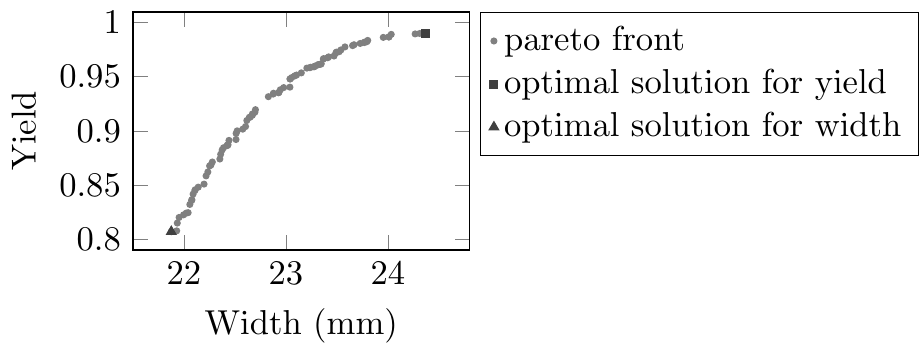}}
	\caption{Single- and multi-objective optimization.}
	\label{fig:YieldOpt}
\end{figure}

For the single-objective optimization in Eq.~\eqref{eq:OptProb} we use the same settings as above and additionally the starting point
$\overline{\up}^0 = [9, 5, 1, 1]^{\transpose}$
and the initial sampling size $\Nmc^{\text{start}}=100$ in the adaptive method. Figures~\ref{fig:YieldOpt_SingleEff} and \ref{fig:YieldOpt_SingleProg} show the increase of the yield and the sampling size in each iteration for the adaptive Newton-MC compared with a classic Newton method with a fixed sample size. Both methods achieve the same optimal yield of $\tilde{Y}_{\text{GPR-H}}(\overline{\up}^{\text{opt}}) = 98.32\,\%$. 
The adaptive Newton-MC needs $11$ iterations and $571$ \gls{fem} evaluations for that, the classic Newton method $37$ iterations and $2643$ \gls{fem} evaluations.

For the multi-objective optimization in Eq.~\eqref{eq:MultiOptFormulation} the python package pymoo has been used, cf.~\cite{pymoo}. We formulate a second objective function to minimize the width of the waveguide. Further we request the yield to be larger than a minimal value $Y_{\text{min}}=0.8$, include this as a constraint and define lower and upper bounds for the mean value of the uncertain parameter. We obtain the optimization problem 
\begin{align}
\min_{\overline{\up}} & \left(-Y(\overline{\up})\right) \label{eq:MultiOpt_WG}\\
\min_{\overline{\up}} & (2p_1+p_2) \notag\\
\quad \text{s.t. } & g_1(\overline{\up}) = Y_{\text{min}}-Y(\overline{\up}) \leq 0, \notag \\
& \left[5,3,0.5,0.5\right]^{\transpose} \leq \overline{\up} \leq \left[25,15,1.5,1.5\right]^{\transpose}. \notag
\end{align}
In pymoo, the NSGA2 solver and the default settings have been used, except for the following modifications: initial population size was set to $200$, number of offsprings per generation to $100$ and maximum number of generations to $30$.
In Fig.~\ref{fig:YieldOpt_MultiFront} we see the pareto front after $30$ generations. Depending on the rating of the significance of the two objective functions, a solution can be chosen.


%

\conclusions
Reliable and efficient methods for yield estimation and optimization have been presented. The hybrid approach based on a \gls{gpr} surrogate model including opportunity to model updates reduces the computational effort significantly, while maintaining high accuracy.  
The proposed adaptive Newton-MC reduces the uncertainty impact with low costs compared to classic Newton methods.
A new multi-objective approach allows optimizing performance and robustness simultaneously.
In future, we will work on improving the efficiency of multi-objective optimization based on genetic algorithms by using adaptive sample size increase and \gls{gpr} approximations of the yield, in addition to the \gls{gpr} approximations of the \glspl{qoi}.










\noappendix       










\begin{acknowledgements}
The work of Mona Fuhrländer is supported by the Graduate School CE within the Centre for Computational Engineering at Technische Universität Darmstadt.
\end{acknowledgements}

\end{document}